\pdfoutput=1
\documentclass[11pt]{article}

\usepackage[]{acl}

\usepackage{times}
\usepackage{latexsym}

\usepackage[T1]{fontenc}
\usepackage[utf8]{inputenc}

\usepackage{microtype}

\usepackage{booktabs}
\usepackage{amsthm}
\usepackage{amsmath}
\usepackage{amssymb}
\usepackage{mathrsfs}
\usepackage{multicol}
\usepackage{multirow}
\usepackage{xspace}
\usepackage{float}
\usepackage{makecell}
\usepackage{tabularx}
\usepackage{graphicx}

\let\oldhat\hat

\renewcommand{\hat}[1]{\oldhat{\mathbf{#1}}}

\newcommand{\ie}{\emph{i.e., }}
\newcommand{\eg}{\emph{e.g., }}

\title{Personalized Topic Selection Model for Topic-Grounded Dialogue}

\author{
Shixuan Fan\textsuperscript{1,2}, Wei Wei\thanks{ \quad  Corresponding author.}\textsuperscript{\ \ 1,2}, Xiaofei Wen\textsuperscript{1,2}, \\
\textbf{Xianling Mao}\textsuperscript{\textbf{3}}\textbf{, Jixiong Chen}\textsuperscript{\textbf{4}}\textbf{, Dangyang Chen}\textsuperscript{\textbf{$\ast$\ 2,5}}\\
\textsuperscript{1}Cognitive Computing and Intelligent Information Processing (CCIIP) Laboratory, \\ School of Computer Science and Technology, Huazhong University of Science and Technology\\
\textsuperscript{2}Joint Laboratory of HUST and Pingan Property \& Casualty Research (HPL)\\
\textsuperscript{3}Department of Computer Science and Technology, Beijing Institute of Technology\\
\textsuperscript{4}Brilliance Technology Co. Ltd. \\
\textsuperscript{5}Ping An Property \& Casualty Insurance company of China\\
\texttt{fanshixuan, weiw, xfwen\}@hust.edu.cn}, \\
\texttt{maoxl@bit.edu.cn}, 
\texttt{chenjixiong@brilliance.com.cn}, \\
\texttt{chendangyang273@pingan.com.cn}
}

\begin{document}
\maketitle

\begin{abstract}
Recently, the topic-grounded dialogue (TGD) system has become increasingly popular as its powerful capability to actively guide users to accomplish specific tasks through topic-guided conversations. Most existing works utilize side information (\eg topics or personas) in isolation to enhance the topic selection ability. However, due to disregarding the noise within these auxiliary information sources and their mutual influence, current models tend to predict user-uninteresting and contextually irrelevant topics. To build user-engaging and coherent dialogue agent, we propose a \textbf{P}ersonalized topic s\textbf{E}lection model for \textbf{T}opic-grounded \textbf{D}ialogue, named \textbf{PETD}, which takes account of the interaction of side information to selectively aggregate such information for more accurately predicting subsequent topics. Specifically, we evaluate the correlation between global topics and personas and selectively incorporate the global topics aligned with user personas. Furthermore, we propose a contrastive learning based persona selector to filter out irrelevant personas under the constraint of lacking pertinent persona annotations. Throughout the selection and generation, diverse relevant side information is considered. Extensive experiments demonstrate that our proposed method can generate engaging and diverse responses, outperforming state-of-the-art baselines across various evaluation metrics.
\end{abstract} 

\maketitle

% Below is a paragraph from an academic paper. Polish the writing to meet the academic style, improve the spelling, grammar, clarity, concision and overall readability. When necessary, rewrite the whole sentence. Furthermore, list all modifications and explain the reasons to do so in markdown table. Paragraph : 

\section{Introduction}

The dialogue systems have been widely used for task-specific interactions like customer service~\cite{zhao2023towards, li2023trea} or emotional interaction~\cite{wei2019emotion, lai2023conversational}. Early works adopt a passive stance in response to user queries, yielding generic or irrelevant responses~\cite{liu2022incorporating,liu2022improving,lu2023miracle}. However, real-world conversational scenarios often manifest heightened complexity, necessitating dialogue agents to adeptly manage topics and actively guide conversations. Although large language models, such as ChatGPT, exhibit closely resemble abilities of humans, they fall short in topic management and exhibit suboptimal initiative~\cite{cao2023diaggpt, hudevcek2023large}. Consequently, topic-grounded dialogue systems (TGDs), which can proactively predict appropriate future topics and generate diverse and informative responses around new topics, have recently attracted considerable attention. 

% 讲了下knowledge-based方案，global-topic based方案，user-based方案和他们的不足之处
Indeed, the core of topic-grounded dialogue systems lies in the effective exploitation of diverse side information (\ie global topics or user personas) to precisely predict subsequent topics. The former~\cite{DBLP:conf/acl/0027L0NWC20, DBLP:conf/emnlp/ZouLHZ21, DBLP:conf/emnlp/Wen0M22} fully models the topic transitions overall topic sequences and infers the subsequent topic via the co-occurrences of adjacent topics over such topic transferring information, and the latter~\cite{DBLP:conf/coling/ZhouZZWW20, DBLP:conf/sigir/RenTLR0XLRC22} takes account of all user personas to model user preferences for topic selection. 

\begin{figure}[t]
  \centering
  \includegraphics[width=\columnwidth]{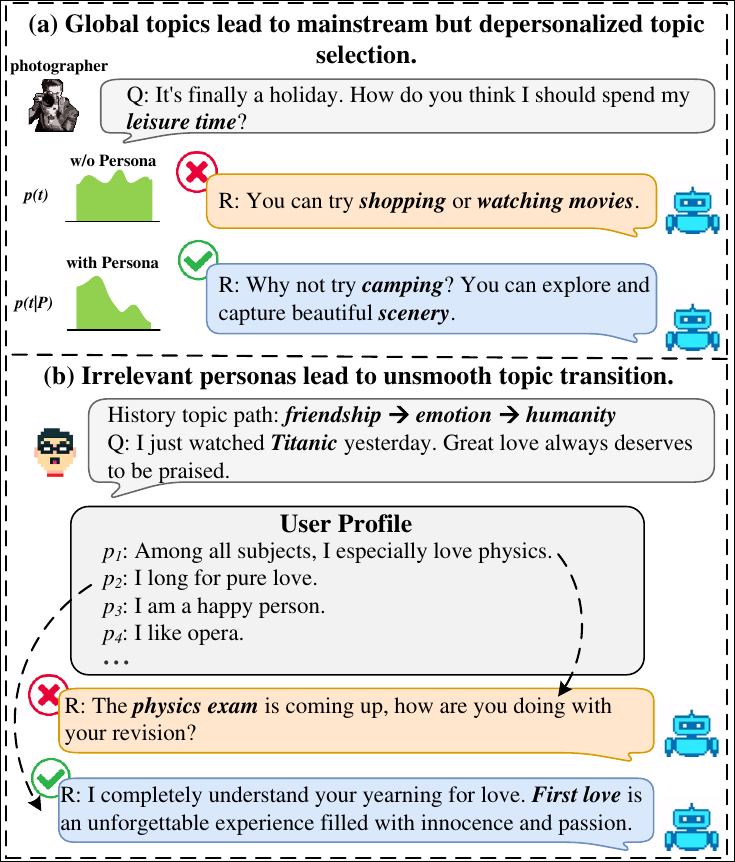}
  \caption{The irrelevant information in global topics and user personas misleads the model to choose depersonalized and unsmooth topics. The topics are bolded and italicized.}
  \label{fig:intr_example}
\end{figure}

% 主要在讲存在的问题和例子
% 1、不同角色的用户可能会根据相同的历史主题选择不同的主题。全局主题频率分布是平坦的因为是多个不同个性用户选择主题的叠加，可能引导模型选择主流但去个性化的主题。
% 2、用户profile包含着多个个性。如图一（b）所示，模型可能被上下文无关的个性误导，选择了不平滑的个性。
% 此外，没有带注释的标签来指示个性和主题间的对应关系，为建模辅助信息间的相互作用和过滤掉其中的无关信息带来了额外的挑战。
Despite achieving promising results, previous studies still face challenges in adequately modeling side information. This limitation arises from inadequate consideration of the interconnections between different side information, leading to the indiscriminate integration of both relevant and irrelevant information for topic selection. This issue can be delineated from two perspectives. \emph{\textbf{Firstly}}, users with different personas may select different topics based on the same history topics. As shown in Figure \ref{fig:intr_example} (a), global co-occurrence topic frequency distribution exhibits a uniform profile due to its amalgamation of diverse persona users choosing topics. Existing works indiscriminately aggregate global topics resulting in biasing models towards mainstream but depersonalized topics. \emph{\textbf{Secondly}}, each selected topic may exclusively align with specific user personas, rather than all of them. As shown in Figure \ref{fig:intr_example} (b), existing methods indiscriminately encode multiple personas in user profiles, resulting in models that may be misled by contextually irrelevant personas and choose unsmooth topics. Furthermore, the absence of annotated labels indicating the correspondence between personas and topics introduces additional challenges for modeling the interaction among side information and filtering out irrelevant information. 

% 主要讲方案
To address these problems, we propose a novel method called \textbf{P}ersonalized topic s\textbf{E}lection model for \textbf{T}opic-grounded \textbf{D}ialogue (\textbf{PETD}), which effectively mitigates the impact of irrelevant side information by considering the interaction among topics and personas for more accurately predicting subsequent topics. To disentangle the global topics that emerge from overlapping numerous personas, we establish the n-to-n correspondence between topics and personas at global level. We further selectively aggregate global topics aligned with user personas for efficiently filtering out irrelevant global topics to user personas and more accurate user preference modeling. To prevent the intrusion of context-irrelevant personas during topic selection, we employ a persona selector to predict personas likely to be manifested by the user in the next turn based on the topic path and dialogue context. In light of existing research, which indicates challenges in ensuring optimal training for each submodule through end-to-end single-task supervised training alone~\cite{zhang2021survey}, we design a contrastive learning based auxiliary task to address the impediment of lacking relevant persona annotations for targeted training of the persona selector submodule. This task aims to specifically optimize the persona selector and amplify the representation distinction between different personas. Finally, we utilize side information after filtering out irrelevant information to assist topic selection and dialogue generation.

To summarize, the contributions of this paper are listed as follows:
\begin{itemize}
    \item We identify the problem in current paradigms, which is the isolated and indiscriminate integration of side information, resulting in depersonalized and unsmooth topic selection. 
    \item We propose a persona-specific global topic expansion method to selectively aggregate global topics under different user personas. 
    \item We exploit fine-grained personas to guide topic selection and a contrastive learning based auxiliary task is proposed to optimize the persona selector and enhance the distinction of different personality representations. 
    % \item Experimental results on a benchmark dataset demonstrate that our proposed PETD can consistently outperform the state-of-the-art baselines in terms of Objective and subjective evaluation metrics.
\end{itemize}

% 作为case的example1
% user profilie:
% 	我非常喜欢有名气的导演
% 	我非常喜欢听笑话
% 	我是一个开心的人
% 	我很喜欢音乐
% 	我喜欢喜剧
% 	我是一个快乐的人
% 	我很喜欢小孩子
% 	我很喜欢明星
% 	想要好运
% 	我很喜欢爱情
% contexts:
% 	0 我们来聊会天吧！	
% 	1 好啊，从小到大，你有没有对你印象特别深刻的老师？	谈论 老师
% 	0 我非常好运，从小到大，每个老师都对我非常好。他们每一个都让我印象深刻。	谈论 好运
% 	1 那真好，不过我上学的时候还没有谈过恋爱，毫无爱情经验，有没有关于爱情的电影，让我学习学习？	谈论 爱情
% 	0 我认为<movie>特别适合你，这是一部苏联爱情电影，非常有看点。	
% 	1 这部我看过，很一般的爱情片，当时我找的那是一个费劲。不过剧情还不错，看完之后我脸上带着微笑。	谈论 微笑
% 	0 是啊，剧情让人回忆起甜甜的初恋。	谈论 初恋
% 	1 我还没有体会过初恋，有没有关于初恋的爱情片？	请求推荐 爱情 请求推荐 初恋
% 	0 你可以看看<movie>，这是一部小清新的公路爱情电影，拍的挺不错的。	
% 	1 看评价是一部清新，轻松，生活化的电影，找时间看一下，应该会很让人快乐。	谈论 快乐
% 	0 恩，的确是一部让人快乐的影片。我再给你推荐一部关于爱情的电影啊。	谈论 爱情
% 	1 好啊，我喜欢纯纯的爱情，还有没有这种电影？	允许推荐 爱情
% 	0 我觉得<movie>不错，里面我喜欢的吴奇隆真帅，是我的白马王子啊！	
% 	1 这个，我看过！电视上看的，好像都是小学的事了。谢谢你的推荐，再见，	

% case 2:
% personas:
% 我非常喜欢听笑话
% 我喜欢美好品德带给我的荣耀
% 我很喜欢喜剧
% 我非常喜欢海军
% 我非常喜欢朝气的少年
% 我喜欢爱情
% 我很喜欢侦探的故事
% 我喜欢柯南
% 我非常喜欢想象力充足的编剧
% 想要好运
% context:
% 最近怎么样？
% 还好吧，我刚调到了我们公司的创意部，这个部门比较清闲，很少加班，现在挺自在的。
% 那挺好啊，哪天有时间我们一起去电玩城疯一天去咋样？
% 我最近也在打电玩，一个以校园爱情为北京的，你要一起玩吗？
% gold response: 看的少了，有时会找些喜剧看看，也挺好看的，要不我给你推荐个喜剧电影，咋样？[/s_response]
% topic_his: 创意 电玩 笑话 喜剧
% gold_topic 喜剧
% global：地球，孩子，隐形，沉迷，心动，灵魂，喜剧

\section{Related Work}

% \subsection{Topic-Grounded Dialogue}
% Researchers have explored various approaches to construct dialogue agents that guide the conversation toward the desired topic, known as target-guided topic-grounded dialogue. 
Nowadays, existing methods in TGD often share a paradigm that decomposes the task into two related sub-task, namely topic selection and response generation~\cite{DBLP:conf/aaai/QinYTL20, DBLP:conf/acl/XuWNWCL20, DBLP:conf/acl/0027L0NWC20, DBLP:conf/emnlp/ZouLHZ21}. 
In this work, we mainly focus on the topic selection task, which can be broadly categorized into knowledge-based, global topic-based and user-based methods. 

% knowledge-based
\noindent \textbf{Knowledge-Based Methods.}
\citet{DBLP:conf/acl/TangZXLXH19} first propose the target-guided dialogue topic selection task and develop a rule-based model based on the similarity of the next topic and target. 
DKRN~\cite{DBLP:conf/aaai/QinYTL20} further utilizes semantic correlation to improve the smoothness of selected topics. 
\citet{DBLP:conf/aaai/XuWNWC20} and CKC~\cite{DBLP:conf/aaai/ZhongLWM21} model topic transition as a continuous walk on the commonsense graph (CKG), which effectively reduces the neighborhood candidate topic space. 
Considering topic selection based on neighborhood entities on CKG does not conform to real dialogues, ECCF~\cite{DBLP:conf/eacl/LiJSYQZZY23} mines high-frequency topic transition from real dialogues to expand commonsense graph. 
However, strict neighborhood constraints in topic selection limit the generalization of these methods to real-world conversations~\cite{li-etal-2022-c3kg}.

% global based
\noindent \textbf{Global Topic-Based Methods.}
This kind of method aims to leverage relevant topic transitions in other paths to aid subsequent topic selection in the local topic path. \citet{DBLP:conf/acl/XuWNWCL20} construct a global topic transition graph utilizing all topic paths in dialogue corpora to capture utterance-level correlations for topic selection. On this basis, ~\citet{DBLP:conf/acl/0027L0NWC20} extend the method by employing Discrete Variational Auto-Encoder (VAE) with Graph Neural Network (GNN) to aggregate global neighborhood topics into local topic paths for modeling global level topic correlations. CG-nAR~\cite{DBLP:conf/emnlp/ZouLHZ21} considers the joint influence of multiple turns of historical topics on the next topic selection and uses a dynamic graph attention mechanism to select subsequent topics. SGTA~\cite{DBLP:conf/emnlp/Wen0M22} represented the global co-occurrence frequency of topics as a multivariate skew-normal distribution with hybrid kernel functions to assist in selecting relevant topics with high global co-occurrence frequencies. 

However, it is crucial to note that users with different personas tend to choose different topics associated with the same historical topic sequence. The aforementioned works indiscriminately fuse the topic paths of users with different personas at the global level, leading to the erroneous modeling of user preferences and depersonalized topic selection. In contrast, PETD exclusively integrates global topics related to the user personas, avoiding interference from global topics corresponding to irrelevant personas. 

% 讨论profile的工作
\noindent \textbf{User-Based Methods.}
Recently, user-based works attempt to model user characteristics for the enhancement of user satisfaction in predicted topics. TG-ReDial~\cite{DBLP:conf/coling/ZhouZZWW20} proposes a topic-grounded dialogue dataset with user personas and employs pre-trained language models to independently encode historical context and all user personas for topic prediction. UPCR~\cite{DBLP:conf/sigir/RenTLR0XLRC22} uses user embedding instead of text-described personas to model user characteristics and combines with dialogue history for topic selection. However, these methods statically model user personas and inject all user personas into each turn of the conversation, resulting in the risk of generating context-independent topics and responses. In contrast, PETD selectively considers relevant personas based on the historical conversation and topic path, taking into account the dynamic transition of the personas exhibited by the user during the conversation.

% \subsection{Personalized Dialogue System}
% The personalized dialogue task aims to require dialogue agents to generate personality-consistent dialogues~\cite{DBLP:conf/ijcai/SongZCWL19, DBLP:conf/emnlp/ChanLYCHZY19, DBLP:conf/sigir/MaDZZW21}. 
% Existing works mainly utilize key-value based~\cite{DBLP:conf/aaai/ZhengZHM20, DBLP:conf/ijcai/QianHZXZ18}, text-based~\cite{DBLP:conf/acl/KielaWZDUS18, DBLP:conf/cikm/Liu0LMFC22} external personalities or implicit persona in history~\cite{DBLP:conf/sigir/MaDZZW21, DBLP:conf/acl/TangWFZHHH23} to improve the diversity and proactivity of generated dialogues. 
% Different from the above works, we focus on using text-based external personalities, which include more comprehensive user preferences, to improve user satisfaction with the topics chosen by dialogue agents. 
% Due to the redundancy of the user personality set, we further model the user's short-term preference based on the current dialogue, which is used to select the fine-grained relevant personas for the upcoming topics.

\section{Method}

\begin{figure*}[t]
  \includegraphics[width=\linewidth]{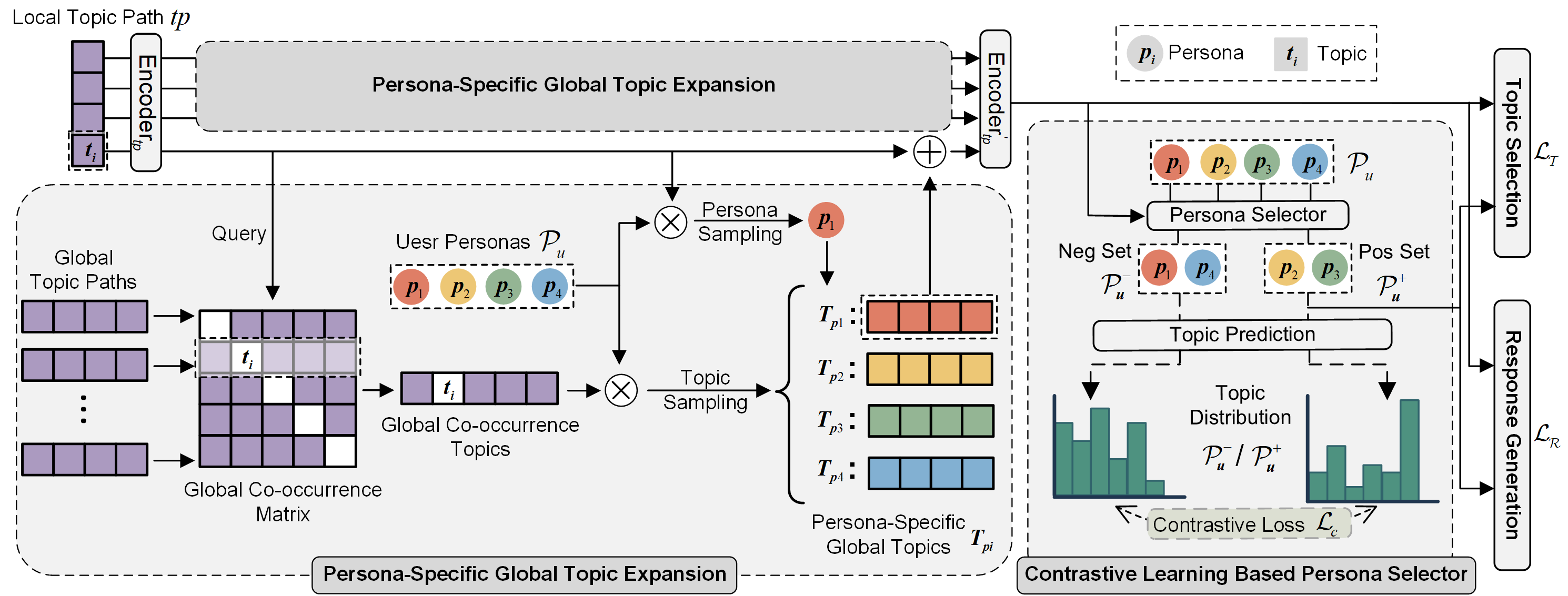}
  \caption{The structure of PETD. We use different colors to represent different personas and their corresponding global topics.}
  \label{fig:main}
\end{figure*}

\subsection{Problem Formulation}

Let $\mathcal{C}=\{c_1, c_2, ..., c_{|\mathcal{C}|}\}$ represent a multi-turn dialogue context and $tp=\{t_1, t_2, ..., t_{|tp|}\}$ is topic path of the dialog $\mathcal{C}$, where $t_j$ refers to the topics discussed at the $j$-th turn. 
We assume that user $u$ is taken from a set $\mathcal{U}$ with a set of predefined personas $\mathcal{P}_u=\{p_1, p_2, ..., p_{|\mathcal{P}_u|}\}$, where each persona is described as a sentence $p_i = \{w_j\}^{|p_i|}_{j=1}$. 
Given conversation $\mathcal{C}_{[1:j]}$, user $u$, user personas $\mathcal{P}_u$ and topic path $tp_{[1:j]}$, the goal of topic-grounded dialogue is to predict topics of the next turn $t_{j+1}$ and generate the corresponding response $\mathcal{R}_{j+1}$. 
We split the target function of the model through the Bayesian formula to correspond to the three subtasks, persona selection $\mathcal{P}^+_{j+1}$, topic selection $t_{j+1}$, and response generation $\mathcal{R}_{j+1}$, separately. 
We formulate target function $y^*$ as follows: 
\begin{equation}
\begin{aligned}
    y^* \triangleq & \prod_{j=1}^{|\mathcal{C}|-1}P(\mathcal{P}^+_{j+1}|u, \mathcal{C}_{[1:j]}, tp_{[1:j]}, \mathcal{P}_u) \cdot \\
                   & \prod_{j=1}^{|\mathcal{C}|-1}P(t_{j+1}|u, \mathcal{C}_{[1:j]}, tp_{[1:j]}, \mathcal{P}^+_{j+1}) \cdot \\
                   & \prod_{j=1}^{|\mathcal{C}|-1}P(\mathcal{R}_{j+1}|u, \mathcal{C}_{[1:j]}, \mathcal{P}^+_{j+1}, t_{j+1}).
\end{aligned}
\end{equation}

For the sake of brevity, we omit all temporal subscripts below. Table \ref{tabel:glossary} in the Appendix lists the notations used in this paper.

\subsection{Input Representation}
We use two multiple multi-layer transformer encoders as the backbone to encode the dialogue context $\mathcal{C}$ and topic path $tp$ respectively. We concat all sentences into a single paragraph before encoding. 
$H_{\mathcal{C}} \in R^{|\mathcal{C}| \times d}$ and $H_{tp} \in R^{|tp|\times d}$ are used to represent the encoded sentence level dialogue context representation and topic representation, respectively.
\begin{equation}
\begin{aligned}
    H_{\mathcal{C}} = & [h_{c_1}, h_{c_2}, ..., h_{c_{|\mathcal{C}|}}] = {\rm encoder}_{\mathcal{C}}(\mathcal{C}), \\
    H_{tp} = & [h_{t_1}, h_{t_2}, ..., h_{t_{|tp|}}] = {\rm encoder}_{tp}(tp).
\end{aligned}
\end{equation}

Users and personas are parameterized as $E_{\mathcal{U}}$ and $E_{\mathcal{P}}$ respectively by indexing corresponding values from the embedding matrix. To utilize the semantic information in persona description, we use BERT~\cite{kenton2019bert} to encode personas at the sentence level as the initialization embedding. For users, we use random initialization embedding.

\subsection{Persona-Specific Global Topic Expansion}
% 对于主题路径中的每一个主题，我们首先构建了一个全局共现主题和所有个性的对应矩阵，来解耦不同个性下的全局共现主题。
In this section, we use personas to selectively fuse global topics to alleviate the detrimental impact of persona-irrelevant global topics on topic selection. For each topic in the topic path, we first calculate the correlation between global co-occurrence topics and user personas to decouple the global topics under different personas. The correspondence score $s_{ij}$ between persona $p_i$ and topic $t_j$ is calculated as follows:
\begin{equation}
    s_{ij} = e_{p_i} W e_{t_j},
\end{equation}
where $W \in R^{d \times d}$ is a learnable matrix, $e_{p_i}$ and $e_{t_j}$ are the embeddings of persona $p_i$ and topic $t_j$. We sample the most relevant $k$ topics for each persona $p_i$, named persona-specific global topics $T_{p_i}$.

% 随后，我们评估每轮历史对话中相关的用户个性。
Subsequently, we select the relevant user personas for each turn of the historical conversation. Context representation $h_{c_i}$, user embedding $e_{u}$, and topic representation $h_{t_i}$ are used together for correspondence score $s'_{ij}$ calculation:
\begin{equation}
    s'_{ij} = f([h_{c_i}; e_u; h_{t_i}], e_{p_j}),
    \label{equ:s}
\end{equation}
\begin{equation}
    f(h_i, h_j) = {\rm MLP}(h_i) \cdot h_j^{\rm T},
    \label{equ:f}
\end{equation}
where [;] represents the concat operation on multiple elements and T means matrix transpose. 
Here, we build a mask matrix $M' \in R^{|tp| \times |\mathcal{P}_u}$ to sample relevant personas. A simple threshold sampling strategy is used to populate the mask matrix according to the correlation score, as follows:
\begin{equation}
    m'_{ij} = 1\ {\rm if}\ \sigma(s'_{ij}) \geq 0.5\ {\rm else}\ 0,
    \label{equ:m}
\end{equation}
where $\sigma$ is the sigmoid activation function. 

% 我们为主题序列融入个性特定的全局主题以得到全局增强后的主题表示{h}'_{t_i}。
For each turn, we aggregate persona-specific global topics corresponding to the selected relevant personas into local topic to obtain global-enhanced topic representation ${h}'_{t_i}$:
\begin{equation}
    h'_{t_i} = {\rm FFN}(h_{t_i} + \sum^{|\mathcal{P}_u|}_{j=1}(s'_{ij} \cdot m'_{ij} \cdot \sum_{t_k \in T_{p_i}}(s_{jk} \cdot e_{t_k}))).
\end{equation}

Another multi-layer transformer encoder is used to encode the topic path after global topic aggregation. 
\begin{equation}
\begin{aligned}
    \widetilde{H}'_{tp} & = {\rm encoder}'_{tp}(H'_{tp})  \\
                        & = {\rm encoder}'_{tp}([h'_{t_1}, h'_{t_2}, ...,h'_{t_{|tp|}}]).
\end{aligned}
\end{equation}

\subsection{Contrastive Learning Based Persona Selector}
In this section, we use a persona selector to select personas that are relevant to the next target topics. 
To optimize the persona selector without supervisory signals and amplify the difference between relevant and irrelevant persona representations, we design a topic prediction auxiliary task based on contrastive learning. 

We use the scoring functions corresponding to formulas \ref{equ:f} and \ref{equ:m} to evaluate the relevance $S'' \in R^{|\mathcal{P}_u|}$ of each persona and construct a mask vector $M'' \in R^{|\mathcal{P}_u|}$.
\begin{equation}
    s''_{p_i} = f([H_{\mathcal{C}}; e_u; \widetilde{H}'_{tp}], e_{p_i}).
\end{equation}

According to the mask vector $M''$, we divide the user persona set into positive persona set $\mathcal{P}^+_u$ (where $m''_{p_i}$=1) and negative persona set $\mathcal{P}^-_u$ (where $m''_{p_i}$=0). 
We further use contrastive learning to reinforce the difference between positive and negative persona sets. 
We first aggregate the positive and negative persona sets separately to obtain the representations $h_{\mathcal{P}^+_u}$ and $h_{\mathcal{P}^-_u}$: 
\begin{equation}
\begin{aligned}
    h_{\mathcal{P}^+_u} &= \sum_{p_i \in \mathcal{P}^+_u} \frac{{\rm exp}(s''_{p_i})}{\sum_{p_i \in \mathcal{P}^+_u} {\rm exp}(s''_{p_i})} \cdot e_{p_i},  \\
    h_{\mathcal{P}^-_u} &= \sum_{p_i \in \mathcal{P}^-_u} \frac{{\rm exp}(-s''_{p_i})}{\sum_{p_i \in \mathcal{P}^-_u} {\rm exp}(-s''_{p_i})} \cdot e_{p_i},
\end{aligned}
\end{equation}
where each weight in the negative set is transformed into a negative value, thereby amplifying the presence of less relevant personas within the negative set representation. We only use personas to predict the next topic, as follows: 
\begin{equation}
    g(h_{\mathcal{P}^+_u}, e_{t_i}) = \frac{{\rm exp}(h_{\mathcal{P}^+_u} W e_{t_i}^{\rm T})}{\sum_{t_j \in \mathcal{T}}{\rm exp}(h_{\mathcal{P}^+_u} W e_{t_j}^{\rm T})}.
\end{equation}

The objective function for contrastive learning is defined as follows: 
\begin{equation}
    \mathcal{L}_c = - \log(g(h_{\mathcal{P}^+_u}, e_{t_i}) - g(h_{\mathcal{P}^-_u}, e_{t_i})),
\end{equation}
where $e_{t_i}$ is the embedding vector of the target topic. The contrastive learning objective function aims to augment the divergence in prediction scores between positive and negative persona sets for subsequent correct topics. Consistent with our intuition, relevant personas tend to predict correct topics while irrelevant personas tend to predict wrong ones. Through this task, we cleverly transform topic annotations into supervision signals of relevant personas to optimize the persona selector.
% 在这里解释对比学习会导致“预测中缺乏中间立场使其变得二元化且灵活性较差。”
% 值得注意的是，我们的工作着重于对话过程中用户个性的动态性。我们设计的对比学习旨在帮助模型。

\subsection{Topic Selection}
We use the same dot product similarity as formula \ref{equ:f} for topic selection, and the probability of each topic being selected is defined as $p(t_i)$: 
\begin{equation}
    p(t_i) = \frac{{\rm exp}(f([H_{\mathcal{C}}; e_u; \widetilde{H}'_{tp}; h_{\mathcal{P}^+_u}]], e_{t_i}))}{\sum_{t_j \in \mathcal{T}} {\rm exp}(f([H_{\mathcal{C}}; e_u; \widetilde{H}'_{tp}; h_{\mathcal{P}^+_u}]], e_{t_j}))},
\end{equation}
where $H_{\mathcal{C}}$ is the dialogue context representation, $e_u$ is the user embedding, $\widetilde{H}'_{tp}$ is the topic path represention after global topic aggregation, $h_{\mathcal{P}^+_u}$ is the positive persona set representation. 
We set a cross-entropy loss to optimize the parameter:
\begin{equation}
    \mathcal{L}_{\mathcal{T}} = - log(p(t_i)).
\end{equation}

\subsection{Response Generation}
We use the transformer decoder with copy mechanism \cite{DBLP:conf/acl/GuLLL16} to generate responses. At the $\zeta$-step, previously generated response embedding $e_{w_{[1:\zeta-1]}}$ and other auxiliary information are fed into the decoder together to generate the current word. 
\begin{equation}
    h_{\zeta} = {\rm decoder}([H_{\mathcal{C}}; e_u; \widetilde{H}'_{tp}; h_{\mathcal{P}^+_u};e_{w_{[1:\zeta-1]}}]).
\end{equation}

The probability of generating the word $w_{\zeta}$ is the sum of both the generation probability $p_g(w_{\zeta})$ and the copy probability $p_c(w_{\zeta})$ : 
\begin{equation}
\begin{aligned}
    p(w_{\zeta}) &= p_g(w_{\zeta}) + p_c(w_{\zeta}), \\
    p_g(w_{\zeta}) &= f(h_t, e_{w_{\zeta}}), \\
    p_c(w_{\zeta}) &= f(h_t, h_{w_{\zeta}}),\ w_{\zeta} \in \mathcal{C}, t_i,
\end{aligned}
\end{equation}
where we use the encoder output $h_{w_{\zeta}}$ of dialogue context and selected topics for copynet generation.

As same as topic selection, we also use the cross-entropy function to optimize model parameters for a response utterance of length $|\mathcal{R}|$:
\begin{equation}
    \mathcal{L}_{\mathcal{R}}    = - \frac{1}{|\mathcal{R}|} \sum^{|\mathcal{R}|}_{\zeta=1}log(p(w_{\zeta})).
\end{equation}

\section{Experiments}
In this section, we demonstrate that our proposed method achieves the state-of-the-art and the importance of each proposed component through exhaustive experiments. We also give detailed hyperparameter analysis (\ref{app:hyper}), persona selection analysis (\ref{app:persona}), and case study (\ref{app:case}) in the Appendix.
% In this section, we provide empirical results on two widely used datasets to demonstrate the effectiveness of our proposed PETD method. The experiment aims to answer the following research questions:
% \begin{itemize}
%   \item {\textbf{RQ1}}: Compared with the state-of-the-art works, how does PETD perform on various metrics of topic selection and response generation?
%   \item {\textbf{RQ2}}: How do the components in PETD affect the performance of the topic selection?
%   \item {\textbf{RQ3}}: Is PETD sensitive to the hyperparameters (history turns $t$, the number of global topics corresponding to each persona $k$)?
%   \item {\textbf{RQ4}}: Can we give an intuitive example to show how PETD leverages user personas to optimize the topic selection and response generation?
% \end{itemize}

\begin{table}[t]
\small
\centering
    \begin{tabular}{ccc}
    \toprule
    \textbf{Dataset}     & \textbf{TG-ReDial} & \textbf{Persona-Chat} \\ \hline
    dialogue             & 10,000    & 9,935        \\
    utterance            & 129,392   & 147,039      \\
    topic                & 2,571     & 2,409        \\
    persona              & 2,433     & 6,737         \\
    avg personas per user & 10        & 5           \\ \bottomrule
    \end{tabular}
\caption{Statistics of the datasets.}
\label{tabel:dataset}
% \vspace{-0.5cm}
\end{table}

\subsection{Datasets} 
To evaluate the effectiveness of PETD, following previous works~\cite{DBLP:conf/emnlp/Wen0M22, DBLP:conf/sigir/RenTLR0XLRC22, DBLP:conf/emnlp/ZouLHZ21, DBLP:conf/coling/KishinamiASTSI22}, we conduct experiments on two widely used benchmark datasets, \textbf{TG-ReDial}~\cite{DBLP:conf/coling/ZhouZZWW20} and \textbf{Persona-Chat}~\cite{DBLP:conf/acl/KielaWZDUS18}, for target-guided topic-grounded dialogue. 
Table \ref{tabel:dataset} presents the statistics of the datasets. 
We give a more detailed description of datasets in the Appendix \ref{app:dataset}.

\begin{table*}[t]
\centering
\resizebox{\textwidth}{!}{
\small
    \begin{tabular}{ccccccccccccc}
    \toprule
    \multirow{2}{*}{\textbf{Dataset}}   & \multirow{2}{*}{\textbf{Method}} & \multicolumn{3}{c}{\textbf{Topic Selection}}   & \multicolumn{8}{c}{\textbf{Dialogue Generation}}   \\ \cmidrule(lr){3-5}   \cmidrule(lr){6-13}
        &   & \textbf{Hit@1}     & \textbf{Hit@3}     & \textbf{Hit@5}     & \textbf{PPL}       & \textbf{BLEU-1}    & \textbf{BLEU-2}    & \textbf{Distinct-1}    & \textbf{Distinct-2}  & \textbf{Relevance}     & \textbf{Fluency}   & \textbf{Informativeness}   \\ \hline
    \multirow{9}{*}{\rotatebox{90}{\textbf{TG-ReDial}}}
        & DKRN            & 0.402 	& 0.482 	& 0.507     & -         & -         & -         & -             & -        & -         & -             & -          \\
        & CKC             & 0.591   & 0.786     & 0.827     & -         & -         & -         & -             & -        & -         & -             & -       \\
        & ECCF            & 0.601 	& 0.839 	& 0.852     & 34.131    & 0.263     & 0.161     & 0.017         & 0.088    & 1.37 	& 1.26 	    & 1.51           \\
        & CG-nAR          & 0.566 	& 0.764 	& 0.829     & 52.417    & 0.161     & 0.103     & 0.015         & 0.047    & 1.45 	& 1.08 	    & 1.42     \\
        & SGTA            & 0.621 	& 0.852 	& 0.867 	& 21.616 	& 0.301 	& 0.191 	& \underline{0.023} 	    & 0.124    & 1.54 	& \underline{\textbf{1.52}} 	    & 1.53     \\
        & Profile-BERT    & 0.499 	& 0.821 	& 0.834 	& 23.552 	& 0.287 	& 0.117 	& 0.019 	    & 0.090    & 1.14 	& 1.48 	    & 1.19     \\
        & TG-CRS          & 0.613 	& 0.816 	& 0.830 	& \underline{19.223} 	& 0.280 	& 0.173 	& 0.021 	    & 0.094     & 1.55 	& 1.51 	    & 1.37    \\
        & UPCR            & \underline{0.808} 	& \underline{0.883} 	& \underline{0.907} 	& 41.234 	& \underline{0.316} 	& \underline{0.200} 	& 0.022      & \underline{0.132}  & \underline{1.57} 	& 1.47 	    & \underline{1.58} \\ \cmidrule(lr){2-13}
        & PETD            & \textbf{0.837}* 	& \textbf{0.899} 	& \textbf{0.920}* 	& \textbf{17.076}* 	& \textbf{0.351}* 	& \textbf{0.224}* 	& \textbf{0.031}* 	    & \textbf{0.176}*   & \textbf{1.75} 	& 1.51 	    & \textbf{1.65}      \\ \hline
    \multirow{9}{*}{\rotatebox{90}{\textbf{Persona-Chat}}}
        & DKRN            & 0.468 	& 0.515 	& 0.533     & -         & -         & -         & -             & -             & -         & -             & -\\
        & CKC             & 0.583   & 0.733     & 0.773     & -         & -         & -         & -             & -             & -         & -             & -\\
        & ECCF            & 0.634 	& 0.775 	& 0.839     & 35.499    & 0.257     & 0.166     & 0.038         & 0.238         & 1.35      & 1.36      & 1.43 \\
        & CG-nAR          & 0.542 	& 0.592 	& 0.613     & 18.237    & 0.184     & 0.134     & 0.041         & 0.237         & 1.34      & 1.27      & 1.39 \\
        & SGTA            & 0.664 	& \underline{0.857} 	& \underline{0.907} 	& \underline{15.648} 	& \underline{0.327} 	& 0.207 	& \underline{0.047} 	    & \underline{0.241}  & 1.52      & 1.43      & \underline{1.57}       \\
        & Profile-BERT    & 0.502 	& 0.820 	& 0.836 	& 19.938    & 0.269     & 0.170     & 0.031         & 0.228         & 1.19      & 1.47      & 1.28 \\
        & TG-CRS          & 0.657 	& 0.846 	& 0.863 	& 16.269 	& 0.285 	& 0.181 	& 0.037 	    & 0.237         & 1.49      & \underline{1.54}      & 1.37 \\
        & UPCR            & \underline{0.685} 	& 0.855 	& 0.865 	& 26.735 	& 0.311 	& \underline{0.216} 	& 0.016 	    & 0.192  & \underline{1.62}      & 1.45     & 1.52\\ \cmidrule(lr){2-13}
        & PETD            & \textbf{0.727}* 	& \textbf{0.904}* 	& \textbf{0.951}* 	& \textbf{13.462}* 	& \textbf{0.346}* 	& \textbf{0.252}* 	& \textbf{0.061}* 	    & \textbf{0.254}   & \textbf{1.67}   & \textbf{1.58}     & \textbf{1.63}      \\ 
    \bottomrule
    \end{tabular}
}
\caption{The performance of PETD and all baselines. The results of the best baseline and best performance in each column are underlined and in boldface respectively. We do not report dialogue generation results for DKRN and CKC because their methods rank candidate sentences instead of generation. Significant improvements compared to the best baseline are marked with * (t-test, p $\le$ 0.05).}
% \vspace{-0.5cm}
\label{tabel:overall_result}
\end{table*}

\subsection{Baselines} 
In order to demonstrate the effectiveness of PETD, we compare it with three category baselines.

\noindent (1) Knowledge-based Methods. 
\textbf{DKRN}~\cite{DBLP:conf/aaai/QinYTL20} proposes an explicit knowledge-routed method for topic selection. 
\textbf{CKC}~\cite{DBLP:conf/aaai/ZhongLWM21} models topic transition on the commonsense graph and utilizes graph neural network to model the correlation between topics. 
\textbf{ECCF}~\cite{DBLP:conf/eacl/LiJSYQZZY23} augments high-frequency topic transitions into the commonsense graph and aggregates neighborhood information using relation-aware attention.

\noindent (2) Global topic-based Methods. 
\textbf{CG-nAR} \cite{DBLP:conf/emnlp/ZouLHZ21} constructs a topic global transition graph for topic selection, and designs a non-autoregressive topic-grounded insertion transformer decoder for response generation. 
\textbf{SGTA} \cite{DBLP:conf/emnlp/Wen0M22} introduces a latent space for flexibly integrating global topic transition probabilities with sequence topic prediction probabilities. 

\noindent (3)User-based Methods. 
\textbf{Profile-Bert, TG-CRS} \cite{DBLP:conf/coling/ZhouZZWW20} introduces profile information for dialogue topic selection for the first time. \textbf{Profile-Bert} uses a pre-trained language model to encode semantic information in all user personas for topic selection. \textbf{TG-CRS} uses user profile, dialogue context, and topic sequence for topic selection. 
\textbf{UPCR}~\cite{DBLP:conf/sigir/RenTLR0XLRC22} uses user embedding and dialogue history to model long-term and short-term user interests respectively for topic selection.

% For more implementation details about baselines and our method, please refer to the Appendix \ref{app:imple}.
\subsection{Implementation Details} \label{app:imple}
All of the baselines and our method are implemented in PyTorch and trained on RTX 4090 24GB. 
We keep a maximum of 7 turns of historical dialogue for all methods and allocate $k$ = 10 global topics for each persona. 
The embedding size is set to 768 and the L2 regularization weight is 1e-6. 
Throughout the experiments, we use Adam optimizer \cite{DBLP:journals/corr/KingmaB14}. Its initial learning rate is 1e-4 and the batch size is set to 80. 
In order to prevent overfitting, the dropout rate is fixed at 0.1. For all datasets, we split the dataset into training/validation/testing sets. We train the model for up to 100 epochs and early stop the training in advance when the hit@3 and bleu-1 don't improve for 10 consecutive epochs on the validation set. 

\subsection{Evaluation Metric}
\noindent \textbf{Automatic Evaluation.} 
We jointly evaluate the abilities of topic selection and response generation from several different perspectives. 
(1) For topic selection accuracy, following previous works~\cite{DBLP:conf/acl/TangZXLXH19, DBLP:conf/coling/ZhouZZWW20, DBLP:conf/emnlp/Wen0M22}, we adopt \emph{Hit@k} (k=1,3,5) as evaluation metrics. 
(2) We report \emph{perplexity} (\emph{PPL}) ~\cite{1995From} and \emph{BLEU-n} (n=1, 2)~\cite{DBLP:conf/acl/PapineniRWZ02} to evaluate the coherence and word overlap of generated utterances. 
(3) Following previous works~\cite{DBLP:conf/coling/ZhouZZWW20, DBLP:conf/emnlp/Wen0M22}, we employ \emph{Distinct-n} (n=1, 2)~\cite{DBLP:conf/naacl/LiGBGD16} to evaluate the diversity of the generated response. 

\noindent \textbf{Human Evaluation.}
Following previous works ~\cite{DBLP:conf/coling/ZhouZZWW20, DBLP:conf/emnlp/Wen0M22}, we adopt \emph{Relevance}, \emph{Fluency} and \emph{Informativeness} of the generated utterances with the rating range of [0, 2]. We recruit three experienced annotators to evaluate 100 randomly selected dialogues. The Fleiss Kappa is 0.68, indicating consistency in the estimates of annotators. The evaluation details are shown in the Appendix \ref{app:human}.

\begin{table*}[t]
\centering
\resizebox{\textwidth}{!}{
\small
    \begin{tabular}{ccccccccccccc}
    \toprule
    \multirow{2}{*}{\textbf{Dataset}}   & \multirow{2}{*}{\textbf{Method}} & \multicolumn{3}{c}{\textbf{Topic Selection}}   & \multicolumn{8}{c}{\textbf{Dialogue Generation}}   \\ \cmidrule(lr){3-5}   \cmidrule(lr){6-13}
        &   & \textbf{Hit@1}     & \textbf{Hit@3}     & \textbf{Hit@5}     & \textbf{PPL}       & \textbf{BLEU-1}    & \textbf{BLEU-2}    & \textbf{Distinct-1}    & \textbf{Distinct-2}  & \textbf{Relevance}     & \textbf{Fluency}   & \textbf{Informativeness}   \\ \hline
    \multirow{4}{*}{\textbf{TG-ReDial}}
        & PETD (0.2B)        & 0.837 & 0.899 & 0.920 & 17.076 & 0.351 & 0.224 & 0.031 & 0.176 & 1.75 & 1.51 & 1.65 \\
        & Llama2 (7B)      & 0.639 & -     & -     & 12.710 & 0.314 & 0.226 & 0.044 & 0.197  & 1.71 & 1.75 & 1.81 \\
        & Llama2-COT (7B)  & 0.658 & -     & -     & 11.492 & 0.349 & 0.239 & 0.042 & 0.193 & 1.78 & 1.74 & 1.81 \\
        & PETD$^{\dagger}$ (7.1B)    & 0.837 & 0.899 & 0.920 & 10.505 & 0.394 & 0.247 & 0.057 & 0.215 & 1.81 & 1.75 & 1.83 \\
        \hline
        
    \multirow{4}{*}{\textbf{Persona-Chat}}
        & PETD (0.2B)        & 0.727 & 0.904 & 0.951 & 13.462 & 0.346 & 0.252 & 0.061 & 0.254 & 1.67 & 1.58 & 1.63 \\
        & Llama2 (7B)      & 0.603 & -     & -     & 11.929 & 0.317 & 0.231 & 0.054 & 0.257 & 1.72 & 1.79 & 1.74 \\
        & Llama2-COT (7B)  & 0.620 & -     & -     & 9.876  & 0.320 & 0.248 & 0.067 & 0.271 & 1.74 & 1.81 & 1.77 \\
        & PETD$^{\dagger}$ (7.1B)    & 0.727 & 0.904 & 0.951  & 9.082 & 0.379 & 0.273 & 0.068 & 0.287 & 1.77 & 1.78 & 1.81 \\
    \bottomrule
    \end{tabular}
}
\caption{The performance of PETD and LLMs. For Llama2 and Llama2-COT, We evaluate the accuracy of topic prediction by detecting whether the target topic is included in the generated responses. We report the number of parameters of each model in symbol ().}
% \vspace{-0.5cm}
\label{tabel:llm}
\end{table*}

\subsection{Main Result}
The evaluation results are shown in Table \ref{tabel:overall_result}.
\subsubsection{Performance on Topic Selection}
Our methods consistently outperform all baselines, achieving an average performance increase of 4.90\%, 3.64\%, and 3.17\% for Hit@1/3/5 respectively compared to the best-performing baseline. This improvement is attributed to PETD, which filters out the irrelevant information from global topics and user persona sets, enabling more precise utilization of side information for dialogue topic selection. 

Overall, global topic-based methods tend to outperform knowledge-based methods due to their capacity to incorporate other topic paths, capturing correlations between different dialogues. Nonetheless, global topic-based methods disregard the influence of persona on topic transition and treat the global information under different personas equally. Conversely, our method differentially considers the global topics under different personas, resulting in improved performance in comparison to global topic-based methods. 

The user-based methods almost achieve the best results overall baselines due to the incorporation of user persona information. It is worth noting that the poor performance of Profile-BERT arises from its failure to dynamically model the personas exhibited by the user during dialogue and using the full personas resulting in noise that heavily interferes with the topic selection. UPCR achieves the best experimental performance among all baselines as it employs user embeddings and topic path to model both long-term and short-term user preferences. However, these user-based methods either consider full personas or solely rely on user embeddings, without adequately tackling the problem of irrelevant information within user persona sets. We design a person selector and optimize it using a contrastive learning-based topic selection auxiliary task to mitigate irrelevant information in user persona sets. In summary, PETD, through its consideration of fine-grained persona, exhibits the ability to provide a more personalized and coherent topic selection compared to state-of-the-art baselines.

% \begin{table*}[!t]
% \centering
% \small
%     \begin{tabular}{cccccccc}
%         \toprule
%         \multirow{2}{*}{\textbf{Method}} & \multicolumn{3}{c}{ \textbf{TG-Redial}} &\multicolumn{3}{c}{ \textbf{Persona-Chat}}   \\ \cmidrule(lr){2-4} \cmidrule(lr){5-7}
%                 & \textbf{Relevance}     & \textbf{Fluency}   & \textbf{Informativeness}   & \textbf{Relevance}     & \textbf{Fluency}   & \textbf{Informativeness}   \\ \hline
%         ECCF            & 1.37 	& 1.26 	    & 1.51      & 1.35      & 1.36      & 1.43              \\
%         CG-nAR          & 1.45 	& 1.08 	    & 1.42      & 1.34      & 1.27      & 1.39              \\
%         SGTA            & 1.54 	& \underline{\textbf{1.52}} 	    & 1.53     & 1.52      & 1.43      & \underline{1.57}              \\
%         Profile-BERT    & 1.14 	& 1.48 	    & 1.19      & 1.19      & 1.47      & 1.28               \\
%         TG-CRS          & 1.55 	& 1.51 	    & 1.37      & 1.49      & \underline{1.54}      & 1.37                \\
%         UPCR            & \underline{1.57} 	& 1.47 	    & \underline{1.58}      & \underline{1.62}      & 1.45     & 1.52            \\ \hline
%         PETD            & \textbf{1.75} 	& 1.51 	    & \textbf{1.65}  & \textbf{1.67}   & \textbf{1.58}     & \textbf{1.63}         \\ 
%         \bottomrule
%     \end{tabular}
% \caption{Performance of PETD and all baselines for human evaluation. 
% % The results of the best baseline and best performance in each column are underlined and in boldface respectively.
% }
% \label{tabel:human}
% \vspace{-0.5cm}
% \end{table*}

\subsubsection{Performance on Response Generation}
Overall, SGTA and UPCR achieve better experimental results in all baselines. The performance of SGTA comes from considering multiple potential topics simultaneously during dialogue generation. The enhanced performance of UPCR results from incorporating user embedding as user characteristics in the generation process. Our method achieves the best performance, with an average improvement of 9.4\%, and 25.9\% in Bleu and Distinct metrics compared to the current state-of-the-art baseline, respectively. The significant improvement in PETD performance arises from a more accurate selection of topics and user personas. The utterances generated by PETD exhibit an average enhancement of 7.28\% and 4.13\% on the relevance and informativeness metrics, respectively, which indicates that our method can better align with the user personas and dialogue topics. 

\subsection{Compaered with LLMs} \label{exp:llm}
% 为了证明主题选择对于Large Language Model的重要性，我们选择了指令微调后Llama2-7b-chat作为backbone进行实验。我们分别采用了两种方式让LLM生成回复，（1）用prompt让模型一次生成回复（Llama2_p）（2）利用思维链让模型选择相关个性后再进行回复生成（Llama2_{COT}）。我们还设计了我们方案的一个增强版变体（PETD$^{\dagger}$），通过用Llama2-7b-chat作为回复生成的解码器。附录描述了更详细的实验细节，实验结果如表所示。
% 我们发现尽管LLM能够生成更加流畅和多样的回复，但是主题选择的准确性比大多数基于主题对话的基线低。这与之前的调研结果相同~\cite{cao2023diaggpt, hudevcek2023large}，在没有外部的主题管理模块时，Llama2与大多数的对话模型一样倾向于讨论当前主题而不是拓展新的主题。相较于Llama2，Llama2(COT)性能的提高说明LLM仍然会受到prompt中无关个性的干扰。PETD$^{\dagger}$在将LLM中引入预测的主题与相关个性后取得了最优的性能，说明我们可以巧妙的利用小模型（主题选择）来提高LLM的主动性和信息性。
To demonstrate the importance of topic selection for the Large Language Models (LLMs), we choose Llama2-7b-chat~\cite{touvron2023llama} after instruction tuning as the backbone for experiments. We use two methods to generate responses for LLM, (1) using the prompt to generate responses for the model at once (Llama2), and (2) using Chain-of-Thought~\cite{wei2022chain} to select relevant personas before generating responses (Llama2-COT). We also design an enhanced variant of our method (PETD$^{\dagger} $) by using Llama2-7b-chat as the decoder to generate responses. For all experiments, we fine-tune three epochs on the corresponding dataset using lora~\cite{hu2021lora} for Llama-7b-chat. The prompts are shown in the Table \ref{tabel:promt} in Appendix.

The experimental results are shown in Table \ref{tabel:llm}. We find that although LLM can generate more fluent and diverse responses, the accuracy of topic selection is lower than most baselines of topic-grounded dialogue. 
This is consistent with previous research results~\cite{cao2023diaggpt, hudevcek2023large}. Without an external topic management module, Llama2, like most conversation models, tends to discuss the current topic rather than expand on new ones. 
Compared to Llama2, the improvement of Llama2-COT indicates that LLM is still subject to interference from irrelevant personas in the prompt. PETD$^{\dagger}$ achieve optimal performance by introducing predicted topics and relevant personas into LLM, indicating that we can cleverly utilize small models (topic selection) to improve the initiative and information of LLM.

\begin{table}[!t]
\centering
\resizebox{\linewidth}{!}{
\small
    \begin{tabular}{cccccc}
        \toprule
        \textbf{Dateset} & \textbf{Method}         & \textbf{Hit@1}     & \textbf{Hit@3}     & \textbf{Hit@5}     \\ \hline
        \multirow{9}{*}{\rotatebox{90}{\textbf{TG-ReDial}}}
        & PETD                    & 0.837 	& 0.899     & 0.920     \\ \cmidrule(lr){2-5}
        & w/o global topic        & 0.801 	& 0.871 	& 0.890     \\
        & w topic similar         & 0.814 	& 0.883 	& 0.910     \\ 
        & w co-occurrence         & 0.813 	& 0.885 	& 0.913     \\ \cmidrule(lr){2-5}
        & w/o persona             & 0.759 	& 0.874 	& 0.894     \\
        & w/o persona selection   & 0.814 	& 0.872 	& 0.902     \\
        & w random persona selection   & 0.729 	& 0.862 	& 0.857     \\
        & w/o auxiliary task      & 0.816 	& 0.877 	& 0.903     \\ 
        & w/o contrastive learning      & 0.823 	& 0.884 	& 0.912     \\  \toprule
        \multirow{9}{*}{\rotatebox{90}{\textbf{Persona-Chat}}}
        & PETD                    & 0.727 	& 0.904     & 0.951     \\ \cmidrule(lr){2-5}
        & w/o global topic        & 0.703 	& 0.834 	& 0.901     \\
        & w topic similar         & 0.704 	& 0.868 	& 0.927     \\ 
        & w co-occurrence         & 0.718 	& 0.875 	& 0.895     \\ \cmidrule(lr){2-5}
        & w/o persona             & 0.676 	& 0.778 	& 0.857     \\
        & w/o persona selection   & 0.718 	& 0.872 	& 0.890     \\
        & w random persona selection   & 0.689 	& 0.798 	& 0.876     \\
        & w/o auxiliary task      & 0.718 	& 0.873 	& 0.893     \\ 
        & w/o contrastive learning      & 0.720 	& 0.887 	& 0.916     \\ 
        \bottomrule
    \end{tabular}
}
\caption{The Performance of Ablation Study.}
\label{tabel:ablation}
% \vspace{-0.5cm}
\end{table}

\subsection{Ablation Study}
To investigate the effectiveness of PETD, we conduct detailed ablation experiments around two key components of PETD. The experimental results are shown in Table \ref{tabel:ablation}. 

\noindent \textbf{Persona-Specific Global Topic Expansion.} We first eliminate the global topic aggregation for w/o global topic. The dramatic drop in this variant performance demonstrates the important role of global information in modeling user preferences. To demonstrate the effectiveness of personas in global topic aggregation, we contrast two variants: one employing topic similarity (w topic similar) and the other utilizing global co-occurrence (w co-occurrence) to select and aggregate global topics. We observe that PETD w co-occurrence and PETD w topic similar achieve similar performance, indicating that the statistical frequency of global co-occurrence can be well fitted by the topic representation similarity. This similarity serves as the basis for our successful decoupling of global topics through personas and topics similarity. However, the performance of these two variants remains notably lower than that of PETD, highlighting the necessity of distinctively accounting for global topics aligned with different personas. 

\noindent \textbf{Contrastive Learning Based Persona Selector.} We design following variants: deletes complete personas (w/o persona), deletes persona selection module (w/o persona selection), instead random selection of persona selection module (w random persona selection), and deletes contrastive learning based auxiliary task (w/o auxiliary task), deletes contrastive learning (w/o contrastive learning), respectively. Generally speaking, richer side information will make the model perform better. When all personas are given (w/o persona selection), the performance of the model is lower than that of selecting personas (PETD), indicating that the model is poisoned by irrelevant persona, proving that persona selection is necessary. For PETD with random persona selection. The number of selected personas is set to 2, consistent with the number of relevant personas selected by our methods (PETD) in most scenarios. Since random selection has a high probability of selecting irrelevant personas, the performance of the PETD w random person selection variant decreases significantly, demonstrating the sensitivity of the model to irrelevant personas. We believe that persona selection is an essential component for intelligent agents as conversation scenarios become more complex and user personas increase significantly. 
We also find that the inclusion or exclusion persona selection module has little impact on model performance when the model is without the auxiliary task. This finding suggests that optimizing the persona selector using a specific end-to-end training method is challenging and the auxiliary task we proposed can subtly address this problem through targeted optimization. After removing contrastive learning, the model's performance drops significantly, although it is slightly higher than the variant that removes all the auxiliary tasks. The significant drop in performance demonstrates the necessity of using contrastive learning to increase the difference between irrelevant and relevant persona representations and to specifically optimize the persona selector.

\section{Conclusion}
In this work, we leverage the interplay between topics and personas to improve the accuracy of topic selection by removing redundant noise in side information. 
We notice that user under different personas selects different topics, and existing global topic-based methods ignore this difference. 
Simultaneously, the complicated persona information in the user profile contains plenty of noise, and only a few are related to the next topic. 
To tackle the above problems, we propose a novel model, named PETD. 
We construct a corresponding topic set for each persona at the global level and selectively amalgamate globally pertinent topic sets aligned with user personas to exclude persona-irrelevant global topics. 
We subsequently develop a persona selector, curbing the adverse influence of irrelevant personas on topic selection. 
A contrastive learning based auxiliary task is proposed to optimize the persona selector and increase the distance between different persona representations in unlabeled scenarios. 
Comprehensive experiments showcase the superior ability of our method to achieve more precise topic selection and produce captivating and varied responses, outperforming all benchmarks across various evaluation metrics.

% 在这篇工作中，我们聚焦于全局主题和用户个性中的噪声问题，通过有选择地融合侧信息有效地提高了主题选择的准确率。我们注意到对话在不同persona的影响下选择的主题不同，传统的基于全局信息的主题选择方案忽视了这一差异性。同时User profile中繁杂的persona信息蕴含大量噪声，只有少数与下一个主题相关。为了解决上述问题，我们提出了一个新的模型，即PETD。我们使用使用persona构建序列内和序列外主题的复杂相关性，从persona角度扩充主题序列以建模topic-set级别的兴趣转移过程。随后，我们设计了一个基于对比学习的主题预测辅助任务，用于进一步优化persona选择，降低噪音对topic选择的干扰。大量的实验证明了我们提出的方案可以生成用户更加感兴趣和多样的回复，在多个评估指标上超越了最好的基线。

\section*{Limitations}
First, our method only uses the simplest top-k sampling and threshold sampling for topic and persona selection and does not experiment with more clever sampling methods. 
Second, due to the large experimental scale and fairness issues, we used a transformer decoder of a similar size to GPT2 in most of the experiments. Although we demonstrated the effectiveness of our method for LLMs in Section \ref{exp:llm}, this is still a limitation. 
Third, considering the simplicity of the method and the gap between the structure of the knowledge graph and real conversations, this study does not discuss how to introduce the knowledge graph into topic prediction. However, external knowledge represented by a knowledge graph can be considered similar to the global topic co-occurrence matrix, as both model topic correlation relationships outside the sequence. Therefore, our method can be extended to other side information such as knowledge graphs through similar decoupling, selection, and fusion methods. 

\section*{Ethics Statement}
In a broad sense, introducing personality information into topic-grounded conversations may indeed lead to user profile privacy leaks and false identity forgery. 
However, in this work, personality information and responses are limited to the scope of the experiment and are not enough to threaten real conversations. 
In addition, all models in this paper are trained and evaluated on datasets collected in the public corpus, and the dataset corpus is only used for experimental purposes. The dataset we use does not contain unethical language.

\section*{Acknowledgments}
This work was supported in part by the National Natural Science Foundation of China under Grant No. 62276110, No. 62172039 and in part by the fund of Joint Laboratory of HUST and Pingan Property \& Casualty Research (HPL). The authors would also like to thank the anonymous reviewers for their comments on improving the quality of this paper. 

% Entries for the entire Anthology, followed by custom entries
\bibliography{anthology,custom}

\clearpage
\appendix

\begin{table}[!t]
\centering
\small
    \begin{tabular}{l p{5.5cm}}                 
        \toprule
        \textbf{Symbol}                  & \textbf{Description}   \\
        \midrule
        $\mathcal{U}$, $u$      & the set of all users and a user   \\
        $\mathcal{P}$, $p$      & the set of all persona and a persona \\
        $\mathcal{T}$, $t$      & the set of all topics and a topic \\
        $\mathcal{C}$, $c$      & a conversation and a utterance in conversation    \\
        $\mathcal{W}$, $w$      & the set of all words and a word    \\
        $\mathcal{R}$           & a response    \\
        $\mathcal{P}_{\mathcal{C}}$ & the persona set of a conversation     \\ 
        $\mathcal{P}^+_{\mathcal{C}}$, $\mathcal{P}^-_{\mathcal{C}}$    & the position and negative persona set of a conversation   \\
        $tp$                    & topic path    \\
        $e$, $E$                & a embedding vector and the embedding matrix \\
        $h$, $H$                & hidden state generated in encoder   \\
        $S$, $s_{ij}$           & correspondence score matrix and the score between $i$ and $j$  \\
        $M$, $m_{ij}$           & mask matrix and the mask value between $i$ and $j$    \\ 
        $W$                     & a learnable parameter matrix  \\
        $d$                     & dimension of embedding and hidden vector  \\
        \bottomrule
    \end{tabular}
\caption{Glossary. }
\label{tabel:glossary}
\end{table}

\section{More Experiments}
\subsection{Datasets} \label{app:dataset}
\noindent \textbf{TG-ReDial}\footnote{https://github.com/Lancelot39/TG-ReDial (Apache-2.0 license)}~\cite{DBLP:conf/coling/ZhouZZWW20} is a dialogue dataset in the movie domain, composed of 10,000 two-party dialogues between a seeker and a recommender. 
The dataset is structured in a topic-guided way, i.e. each dialogue in the TG-ReDial dataset includes a topic path to achieve target-guided topic-grounded dialogue. Each dialogue has 7.9 topics and each utterance contains 19.0 words. 
We use the same data preprocessing strategy as in \cite{DBLP:conf/emnlp/Wen0M22, DBLP:conf/sigir/RenTLR0XLRC22}.

\noindent \textbf{Persona-Chat}\footnote{https://parl.ai/projects/personachat/ (MIT license)}~\cite{DBLP:conf/acl/KielaWZDUS18} is an open-domain dialogue dataset, which covers a broad range of topics. 
Following previous works \cite{DBLP:conf/aaai/QinYTL20, DBLP:conf/aaai/ZhongLWM21, DBLP:conf/emnlp/ZouLHZ21, DBLP:conf/coling/KishinamiASTSI22}, we use TF-IDF and part-of-speech (POS) to extract topics from dialogue utterance. 
Inspired by~\cite{DBLP:conf/aaai/ZhongLWM21, DBLP:conf/coling/KishinamiASTSI22}, we use the first topic in the follow-up dialogue that does not constant movement on the ConceptNet~\cite{DBLP:conf/aaai/SpeerCH17} as the target to construct the dataset into the form of target-guided topic-grounded dialogue task.

\begin{table}[t]
\small
\centering
    \begin{tabular}{p{7.2cm}}
    \toprule
    \textbf{Relevance} \\
    2: Fits the user's personality and is related to the current conversation\\
    1: Relevant to one of the user's personality or current conversation, irrelevant or conflicting with the other  \\
    0: Does not resemble any user's personality and context history \\ 
    \midrule
    \textbf{Fluency} \\
    2: Fluent and easy to read \\ 
    1: Grammatically formed \\  
    0: Not a complete sentence or hard to read \\  
    \midrule
    \textbf{Informativeness} \\ 
    2: Have clear and specific meaning \\
    1: Contain a few informative words \\ 
    0: Meaningless sentence  \\
    \bottomrule
    \end{tabular}
\caption{Criteria of human evaluation.}
\label{table:human}
\end{table}

\subsection{Human Evaluation Metrics} \label{app:human}
\emph{Relevance} is used to evaluate the relevance of selected topics and generated sentences to historical conversations and the user's personality. \emph{Fluency} is used to measure the fluency of generated utterances. \emph{Informativeness} is used to evaluate whether the generated utterance revolves around the topics and user personas. The detailed scoring criteria are shown in Table \ref{table:human}. 

Considering that we conduct evaluations on Chinese and English datasets, each evaluator we recruit is required to be a native Chinese speaker and have a high level of English proficiency. All human evaluations are conducted anonymously.

\begin{table*}[t]
\centering
\small
    \begin{tabular}{c|p{14cm} c}
        \toprule
        Model   & Prompt    \\
        \midrule
        \multirow{10}{*}{\rotatebox{90}{\textbf{Llama2}}}  
            & The following is a conversation between an AI assistant called Assistant and a human user called User. The assistant needs to guide the conversation to target topic based on the user's personality and historical conversations. \\
            & The users' personas are following: \\
            & <persona 1>, <persona 2>, ...,<persona n>    \\
            & The topic history is following: <topic 1>, <topic 2>, ..., <topic n>   \\
            & The target topic is <target topic>    \\
            & The conversation is following: \\
            & Assistant: <utterance 1>, User: <utterance 2>, Assistant: <utterance 3>,  ..., User: <utterance n>,\\
            & Assistant: \\
        \midrule
        \multirow{16}{*}{\rotatebox{90}{\textbf{Llama2(COT)}}}  
            & \textbf{Turn 1:}   \\
            & The following is a conversation between an AI assistant called Assistant and a human user called User. The assistant needs to guide the conversation to target topic based on the user's personality and historical conversations. Please think step by step and first predict the personas that users may be interested in in the next turn.\\
            & The users' personas are following: \\
            & <persona 1>, <persona 2>, ...,<persona n>    \\
            & The topic history is following: <topic 1>, <topic 2>, ..., <topic n>   \\
            & The target topic is <target topic>    \\
            & The conversation is following: \\
            & Assistant: <utterance 1>, User: <utterance 2>, Assistant: <utterance 3>,  ..., User: <utterance n>,\\
            & Relevant Personas:\\
            & \textbf{Turn 2:}   \\
            & Please generate a response based on the selected personas and conversation history, and gradually guide the conversation to the target topics. \\
            & Assistant: \\
        \bottomrule
    \end{tabular}
\caption{The prompt of Llama2 and Llama2(COT). The symbol <> is a placeholder that represents the corresponding data in the dataset.}
\label{tabel:promt}
\end{table*}

\begin{figure}[!t]
    \centering
    \begin{minipage}[t]{0.49\linewidth}
        \centering      \includegraphics[width=1\linewidth]{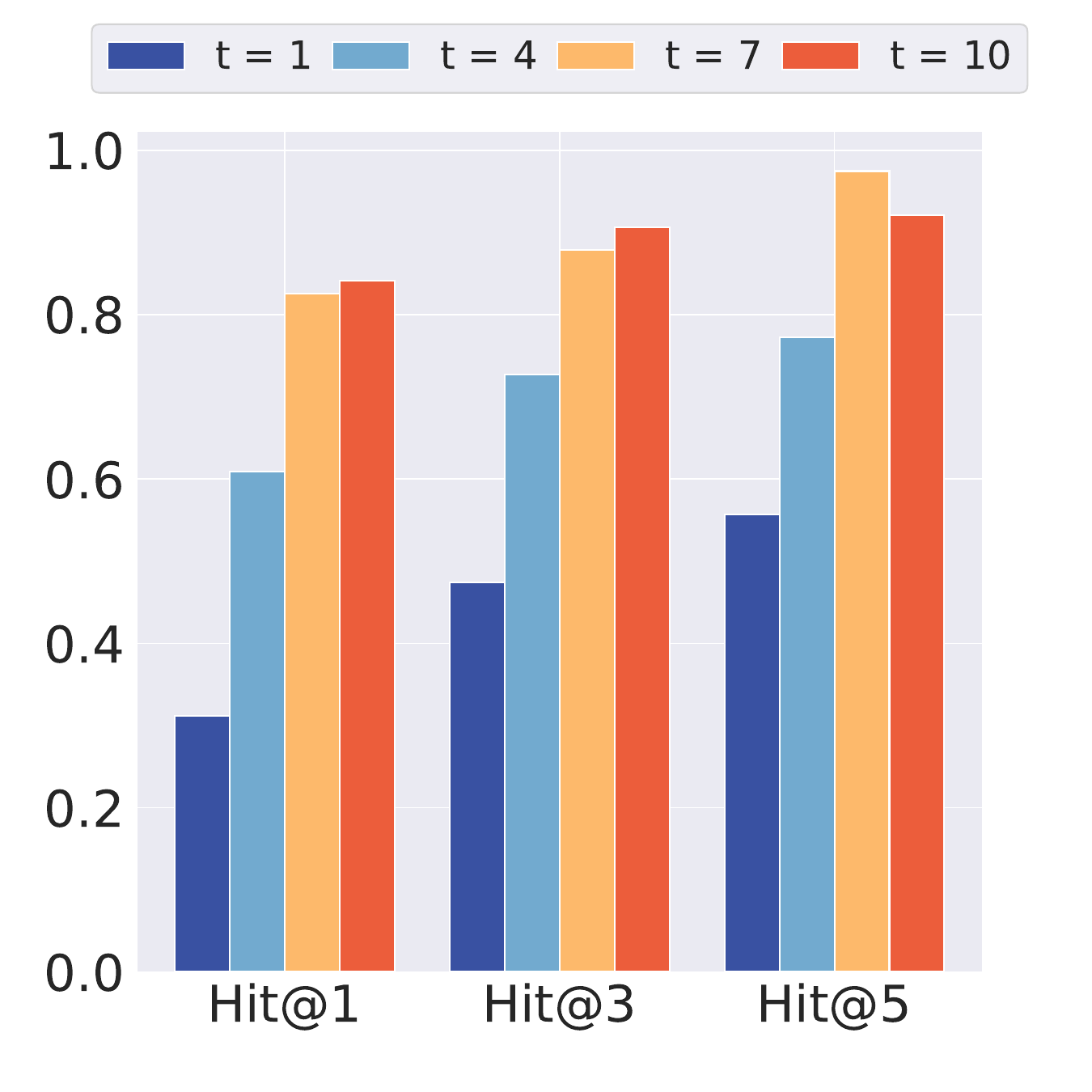}
        \centerline{(a) history turns}
    \end{minipage}
    \begin{minipage}[t]{0.49\linewidth}
        \centering      \includegraphics[width=1\linewidth]{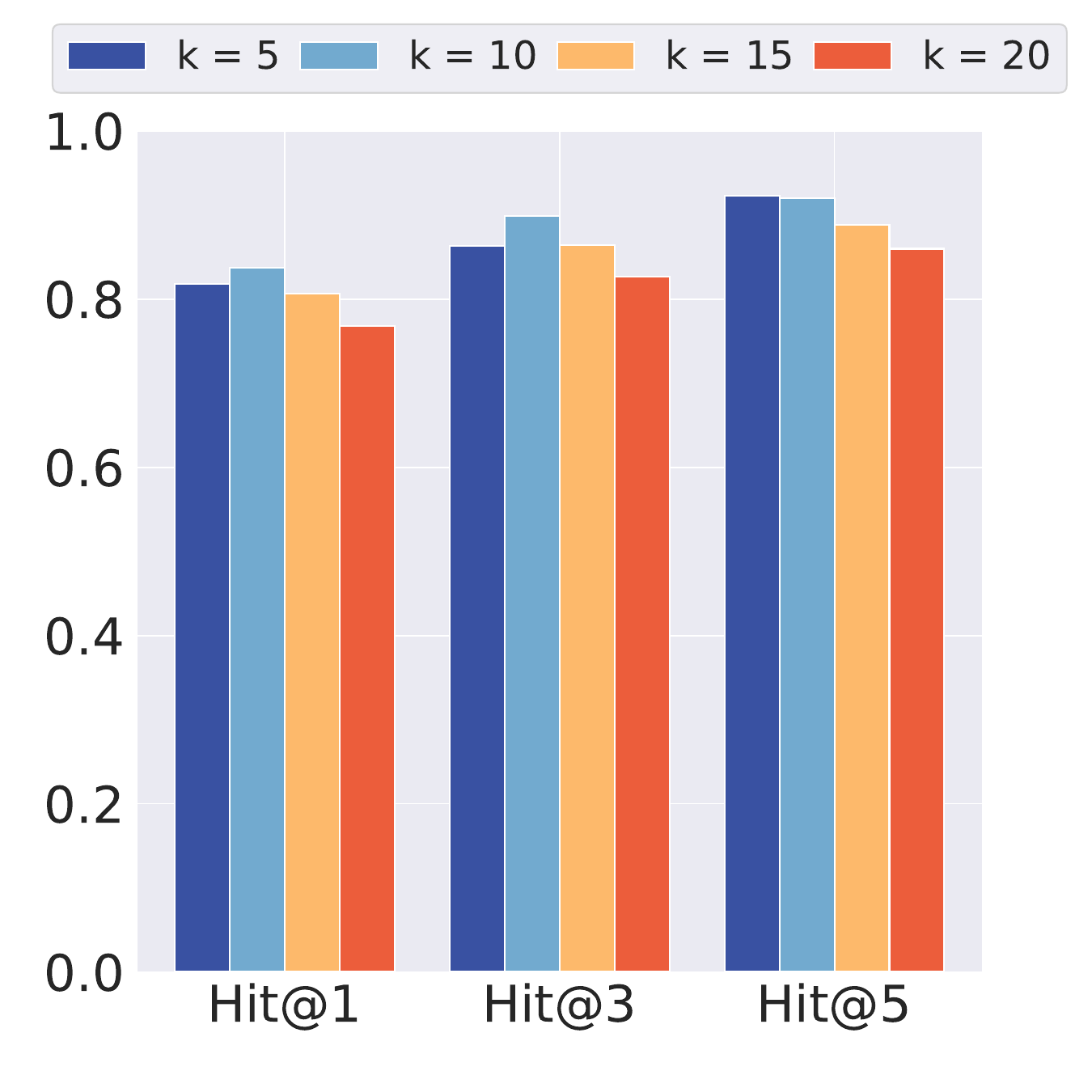}
        \centerline{(b) topics per persona}
    \end{minipage}
    \caption{Impact of the number of history turns (a) and topics per persona (b) on TG-ReDial dataset.}
    \label{figure:hyper}
\end{figure}

\subsection{Hyperparameter Research} \label{app:hyper}
To explore the sensitivity of our proposed PETD model to hyperparameters, we conduct experiments on two hyperparameters, the number of history turns $t \in \{1, 4, 7, 10\}$, global topics corresponding to each persona $k \in \{5, 10, 15, 20\}$ on TG-ReDial dataset. 
The experimental results are shown in Figure \ref{figure:hyper}. 

The performance of the model increases with the number of history turns, which is consistent with previous works ~\cite{DBLP:conf/emnlp/ZouLHZ21, DBLP:conf/emnlp/Wen0M22}, indicating that the user's next topic is related to long historical dialogue, and it is necessary to model interest transition in historical dialogues. 
We observe that the improvement of model performance slows down significantly when the number of turns reaches 7 rounds. 
To balance efficiency and performance, we choose 7 turns as the hyperparameter of the main experiment. 
The performance of PETD is not sensitive to the number of topics corresponding to each persona, but the performance of the model still degrades when $n$ is too small or large. 
When $k$ exceeds 10, the performance of the model gradually decreases as $k$ increases, indicating that forcibly assigning too many topics for each persona will introduce irrelevant information.

\begin{table}[!t]
\centering
\resizebox{\linewidth}{!}{
\small
    \begin{tabular}{cccccc}
        \toprule
        \textbf{Dateset} & \textbf{Method}         & \textbf{Recall}     & \textbf{Precision}     & \textbf{F1}     \\ \hline
        \multirow{3}{*}{\textbf{TG-ReDial}}
        & PETD                    & 0.754 & 0.854 & 0.801 \\
        & PETD w/o auxiliary task & 0.614 & 0.546 & 0.578 \\
        & Llama2-COT              & 0.415 & 0.472 & 0.442 \\
        \toprule
        \multirow{3}{*}{\textbf{Persona-Chat}}
        & PETD                    & 0.790 & 0.833 & 0.811 \\
        & PETD w/o auxiliary task & 0.674 & 0.508 & 0.579 \\
        & Llama2-COT              & 0.516 & 0.489 & 0.502 \\
        \bottomrule
    \end{tabular}
}
\caption{The performance of persona selection.}
\label{tabel:persona}
\end{table}

\subsection{Persona Selection Analysis} \label{app:persona}
We evaluated the persona prediction accuracy of our proposed method (PETD), a variant of PETD that deletes the contrastive learning based auxiliary task (PETD w/o auxiliary task), and Llama2-7b-chat using the thinking chain (Llama2-COT). To evaluate the accuracy of persona selection, we manually annotated 100 pieces of data with relevant personas for each dataset. We adopt \emph{Recall}, \emph{Precision} and \emph{F1} as the evaluation metrics. The experimental results are shown in Table \ref{tabel:persona}.

We find that our proposed method accurately predicts the relevant personas that users are likely to display in the next turn. 
The PETD w/o auxiliary task variant has a significant decrease in precision, indicating that without contrastive learning based auxiliary task, the model has difficulty distinguishing between relevant and irrelevant personas, and tends to predict more irrelevant personas. 
We found that the reason for the lower performance of Llama2-COT is that, through prompts, Llama2 prefers to select relevant personas shown in the conversation history rather than predicting relevant personalities in the future.

\begin{table*}[h!]
\centering
\small
    \begin{tabular}{c|p{7cm} p{7cm}}
        \toprule
        \multirow{5}{*}{\rotatebox{90}{\textbf{Personas}}}  
            & $p_1$: I like expressing emotions.  & $p_6$: I like music very much.\\
            & $p_2$: I seem to fall in love lately.  & $p_7$: I really like the impressive screenplay.  \\
            & $p_3$: I like children very much.   & $p_8$: I'm currently studying psychology. \\
            & $p_4$: I yearn for love very much.  & $p_9$: I like famous actors.  \\
            & $p_5$: I am a person who is easily moved.   & $p_{10}$: I want good luck. \\
            \midrule
        \multirow{6}{*}{\rotatebox{90}{\textbf{Context}}}    
            & \multicolumn{2}{p{14cm}}{A: How have you been lately?} \\ 
            & \multicolumn{2}{p{14cm}}{B: I've been fine lately. How long does it take for an \emph{\textbf{actor}} to become popular since their debut?}  \\
            & \multicolumn{2}{p{14cm}}{A: There is no specific standard for this, some are very fast and some are many years. Anyway, it is not easy for them to be well known by everyone, and some of them \emph{\textbf{succeed}} by certain means.} \\
            & \multicolumn{2}{p{14cm}}{B: Such success is not worth mentioning, let's talk about something else, are you going to take the exam soon, \emph{\textbf{good luck}} to you.} \\ 
            \midrule
        \multirow{10}{*}{\rotatebox{90}{\textbf{Response}}}      
            & \multicolumn{2}{p{14cm}}{\textbf{Gold Response:} Haha, I hope this time it will pass smoothly. When you say I can reap \emph{\textbf{love}} again, it will be even more perfect.} \\ \cmidrule{2-3}
            & \multicolumn{2}{p{14cm}}{\textbf{SGTA:} Thank you, how lucky I am to have your \emph{\textbf{friendship}}.}   \\ \cmidrule{2-3}
            & \multicolumn{2}{p{14cm}}{\textbf{TG-CRS:} Let's not talk about exams, have you listened to \emph{\textbf{music}} recently, Stephen Chow's new song?} \\ \cmidrule{2-3}
            & \multicolumn{2}{p{14cm}}{\textbf{UPCR:} I know a way to improve \emph{\textbf{luck}}, you know?} \\   \cmidrule{2-3}
            & \multicolumn{2}{p{14cm}}{\textbf{Llama2:} Thank you for your blessings. By the way, I have watched many movies recently. Can I recommend some movies about \emph{\textbf{love}} to you?}    \\   \cmidrule{2-3}
            & \multicolumn{2}{p{14cm}}{\textbf{PETD:} Hey, the exam can pass smoothly. I just watched a movie about love last night, and I was moved to tears. I don’t know when I will reap my own \emph{\textbf{love}} again. } \\
            & \textbf{select personas:} $p_2$, $p_4$, $p_5$ \\ 
            \bottomrule
    \end{tabular}
\caption{One case extracted from TG-ReDial. The topics are bolded and italicized.}
\label{tabel:case}
\end{table*}

\begin{figure*}[h!]
\centering
    \begin{minipage}[t]{0.47\linewidth}
        \centering      \includegraphics[width=1\linewidth]{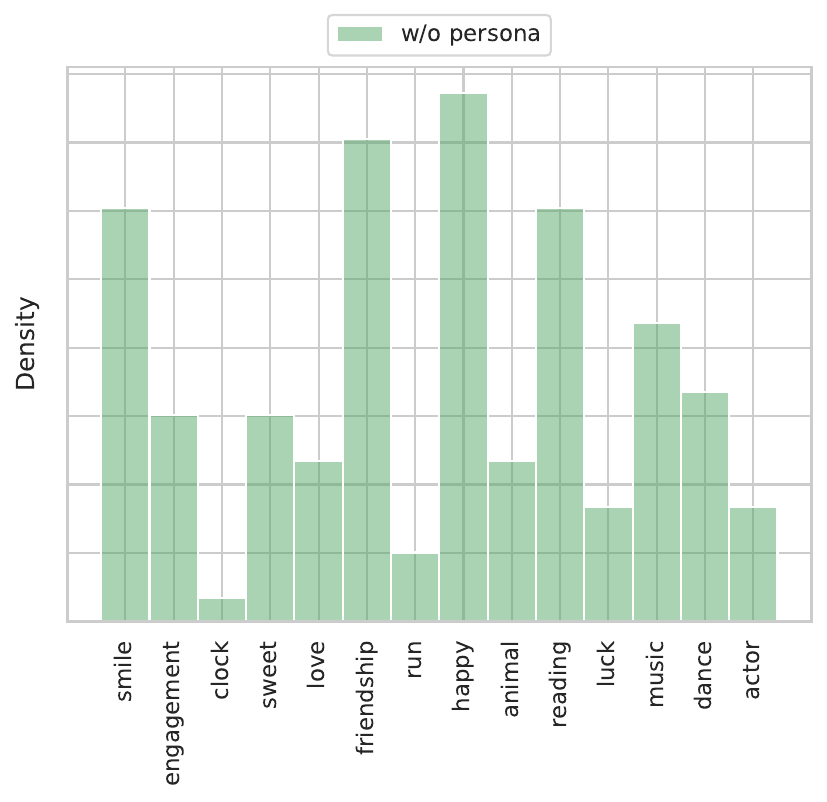}
    \end{minipage}
    \begin{minipage}[t]{0.47\linewidth}
        \centering      \includegraphics[width=1\linewidth]{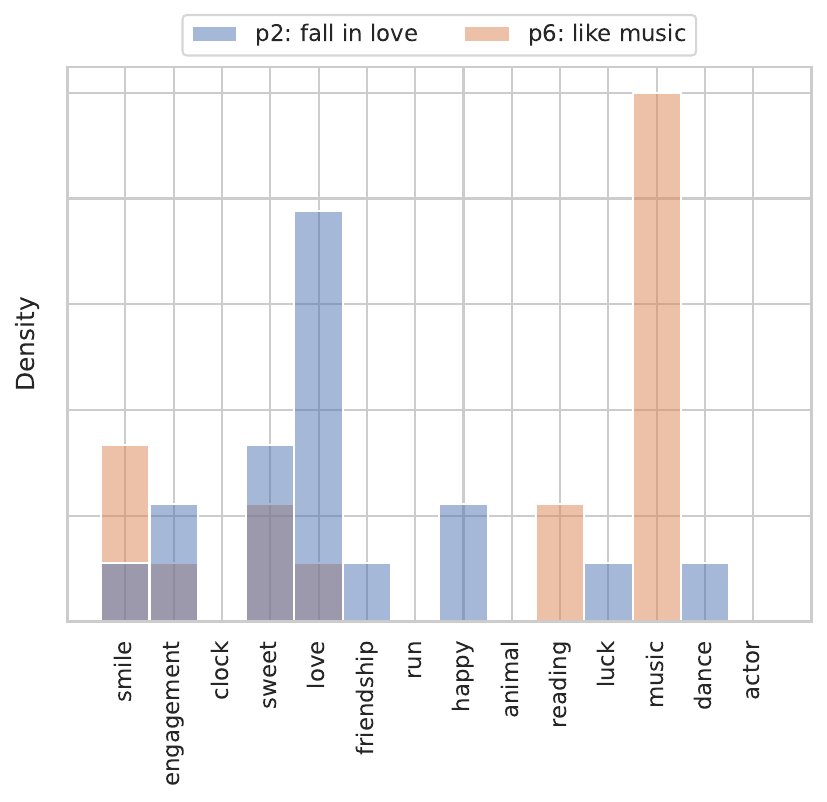}
    \end{minipage}
    \caption{The global co-occurrence frequency density plot of the topic `good luck' in the TG-ReDial dataset. For brevity, we do not draw all global topics.}
    \label{figure:co}
\end{figure*}

\subsection{Case Study} \label{app:case}
We provide an example from the TG-ReDial dataset in Table \ref{tabel:case}. We notice that SGTA chooses globally relevant \emph{friendship} as the topic because of its lack of consideration for user personas. Although TG-CRS considers the persona information, it brutally injects a whole range of personas, relevant and irrelevant, into the topic selection model and chooses persona-relevant but incorrect \emph{music} as the topic. UPCR, which models long-term preference based on user embeddings, mistakenly chooses \emph{luck} as the topic due to the lack of explicit modeling of each persona's fine-grained granularity. In contrast, PETD accurately models user preference, selects the corresponding personas, and further chooses the correct topics by taking into account the interaction of personas and topics to selectively aggregate relevant side information. Additionally, the PETD response is also affected by the persona of `I am a person who is easily moved.', and generated utterance `I was moved to tears', which effectively increased the user's interest in the dialogue. The improved performance of PETD is attributed to filtering out irrelevant information on global topics and user personas.

We further provide the global co-occurrence frequency density plot of the topic `good luck' in the TG-ReDial dataset. As shown in Figure \ref{figure:co}, the co-occurrence frequency density under the global topic and a certain persona has obvious distribution differences. We find that the global co-occurrence topic frequency is a multimodal distribution because it can be regarded as a superposition state of the global co-occurrence topic distribution under multiple personas. This explains that indiscriminately incorporating side information often leads the model to irrelevant or uninteresting topic choices.

% personas:
% 我很喜欢表达情绪
% 我最近有点心动
% 我非常喜欢儿童
% 我非常喜欢爱情
% 我是一个容易被感动感动的人
% 我非常喜欢音乐
% 我非常喜欢好看的剧本
% 我最近在研究心理
% 我很喜欢著名的演员
% 想要好运
% context:
% 最近怎么样。
% 最近还好，一个演员从出道到走红，需要多久？
% 这个可是没有具体标准呢，有点很快有点好多年呢，反正他们要想能成功的被大家熟识都是很不容易的，还有些个是靠某些手段成功的。
% 这样的成功不值得一提，说点别的吧，你是不是快要去考试了，祝你好运吧。
% gold response: 哈哈，但愿这次能顺利通过，你说啥时候我能再收获爱情就更完美了。[/s_response]
% pred response: 哎，考试应该没问题。昨晚刚看了一个关于爱情的电影，我都感动哭了，不知道什么时候才能有属于自己的爱情呀。
% topic_his: 演员 成功 好运 爱情
% gold_topic 爱情

\end{document}